\documentclass[10pt,letterpaper]{article}
\usepackage[top=0.85in,left=2.75in,footskip=0.75in]{geometry}

\usepackage{amsmath,amssymb}

\usepackage{changepage}

\usepackage[utf8x]{inputenc}

\usepackage{textcomp,marvosym}

\usepackage{cite}

\usepackage{nameref,hyperref}

\usepackage{microtype}
\DisableLigatures[f]{encoding = *, family = * }

\usepackage[table]{xcolor}

\usepackage{array}

\definecolor{purple}{rgb}{0.45, 0.31, 0.59}

\newcolumntype{+}{!{\vrule width 2pt}}

\newlength\savedwidth

\usepackage{setspace}
\usepackage{indentfirst} 

\usepackage{color} %

\raggedright
\usepackage[aboveskip=1pt,labelfont=bf,labelsep=period,justification=raggedright,singlelinecheck=off]{caption}

\makeatletter
\renewcommand{\@biblabel}[1]{\quad#1.}
\makeatother

\usepackage{lastpage,fancyhdr,graphicx}
\fancyheadoffset[L]{2.25in}
\fancyfootoffset[L]{2.25in}
\usepackage{bm}
\usepackage{multirow}
\usepackage{here}

\usepackage{color}

\begin{document}
\vspace*{0.2in}

\begin{flushleft}
{\Large
\textbf\newline{Deep learning generates custom-made logistic regression models for explaining how breast cancer subtypes are classified}
}
\newline
\\
Takuma Shibahara\textsuperscript{1\P, \#a, *},
Chisa Wada\textsuperscript{2\P},
Yasuho Yamashita\textsuperscript{1\P},
Kazuhiro Fujita\textsuperscript{2\P},
Masamichi Sato\textsuperscript{2},
Junichi Kuwata\textsuperscript{1},
Atsushi Okamoto\textsuperscript{2},
Yoshimasa Ono\textsuperscript{3}
\\
\bigskip
\textbf{1} Research and Development Group, Hitachi Limited, Tokyo, Japan
\\
\textbf{2} Bioinformatics Group, Translational Research Department, Daiichi Sankyo RD Novare Coporation, Limited, Tokyo, Japan
\\
\textbf{3} Translational Research Department, Daiichi Sankyo RD Novare Coporation, Limited, Tokyo, Japan
\bigskip

\P These authors contributed equally to this work.

\#a Current Address: Central Research Laboratory, Hitachi Limited, Tokyo, Japan

* takuma.shibahara.nj@hitachi.com

\end{flushleft}

\section*{Abstract}
Differentiating the intrinsic subtypes of breast cancer is crucial for deciding the best treatment strategy. Deep learning can predict the subtypes from genetic information more accurately than conventional statistical methods, but to date, deep learning has not been directly utilized to examine which genes are associated with which subtypes. To clarify the mechanisms embedded in the intrinsic subtypes, we developed an explainable deep learning model called a point-wise linear (PWL) model that generates a custom-made logistic regression for each patient. Logistic regression, which is familiar to both physicians and medical informatics researchers, allows us to analyze the importance of the feature variables, and the PWL model harnesses these practical abilities of logistic regression. In this study, we show that analyzing breast cancer subtypes is clinically beneficial for patients and one of the best ways to validate the capability of the PWL model. First, we trained the PWL model with RNA-seq data to predict PAM50 intrinsic subtypes and applied it to the 41/50 genes of PAM50 through the subtype prediction task. Second, we developed a deep enrichment analysis method to reveal the relationships between the PAM50 subtypes and the copy numbers of breast cancer. Our findings showed that the PWL model utilized genes relevant to the cell cycle-related pathways. These preliminary successes in breast cancer subtype analysis demonstrate the potential of our analysis strategy to clarify the mechanisms underlying breast cancer and improve overall clinical outcomes.


\section*{Introduction}
Seven decades after the birth of the learning machine \cite{doi:10.1093/mind/LIX.236.433}, deep learning has evolved to the point that it can provide various predictive analyses. As deep learning spreads into more and more applications, including bioinformatics analysis, there is a growing need to explain the reasons for its predictions. Many methods that evaluate the importance of individual features have been devised to make deep learning models more explainable. These methods can be roughly classified into perturbation-based and saliency-based. In both types, the importance is determined by how much each feature contributes to the output. Perturbation-based methods calculate an importance score based on how the output behaves in relation to a perturbed input \cite{zeiler2014visualizing, zintgraf2017visualizing, ribeiro2016should}. In saliency-based methods, the importance score depends on each feature's saliency evaluated by the gradient of the output with respect to the input \cite{simonyan2013deep,sundararajan2016gradients, bach2015pixel, shrikumar2017learning}.

In the current study, we developed a point-wise linear model (PWL) for innately explainable deep learning in RNA-sequencing (RNA-seq) analysis. Conventional deep learning models compute new feature vectors with a linear combination that sufficiently expresses the objective model, while in contrast, the network of the proposed PWL model derives a weight function for each original feature vector as a function of the original feature vectors. Specifically, it generates a custom-made linear model (e.g., logistic regression) for each sample, and unlike a simple linear model, each linear model it generates involves the nonlinear interactions between the original features, since the weight functions depend on the original feature vector. At the same time, the importance of each feature can be evaluated by its weight function in each linear model. This property is highly advantageous in medical applications because practitioners can utilize the know-how of medical data analysis accumulated throughout its long history. Additionally, when the PWL model is utilized for deep learning that accurately predicts cancer subtypes, it can potentially access unknown and nonstandard knowledge related to gene expressions.

Breast cancer is the most frequently found cancer in women and is the type of cancer most often subjected to genetic analysis. Even so, it is a leading cause of cancer-related deaths in women. Analyzing breast cancer is clinically beneficial for patients and one of the best ways to validate the capability of the PWL model. Conventionally, breast cancer has been classified on the basis of the protein expression of the estrogen receptor (ER), progesterone receptor (PR), and epidermal growth factor receptor ErbB2/Her2, and expressions of these receptors have been used as clinicopathological variables for treatment decisions \cite{reis2011gene}. Since the early 2000s, high-throughput genomics technologies have demonstrated that breast cancer has five clinically relevant molecular subtypes defined by intrinsic gene expression patterns of the cancer \cite{reis2011gene, perou2000molecular, sorlie2003repeated, parker2009supervised, cancer2012comprehensive}: Luminal A, Luminal B, Her2-enriched, basal-like, and normal breast-like cancer. While the subtypes do not perfectly reflect the clinical features, most breast cancers of the luminal subtypes are ER/PR-positive, most Her2-enriched ones have amplification of the Her2 gene, and most basal-like ones are triple negative (ER−/PR−/Her2−). In the original PAM50 study, the classification of the normal breast-like cancer subtype was trained with normal breast tissue \cite{parker2009supervised}. Therefore, cancer samples classified to the normal-like subtype are often interpreted as low tumor content samples \cite{parker2009supervised, weigelt2010breast}.

\begin{figure}[h]
\begin{center}
\includegraphics[width=\linewidth]{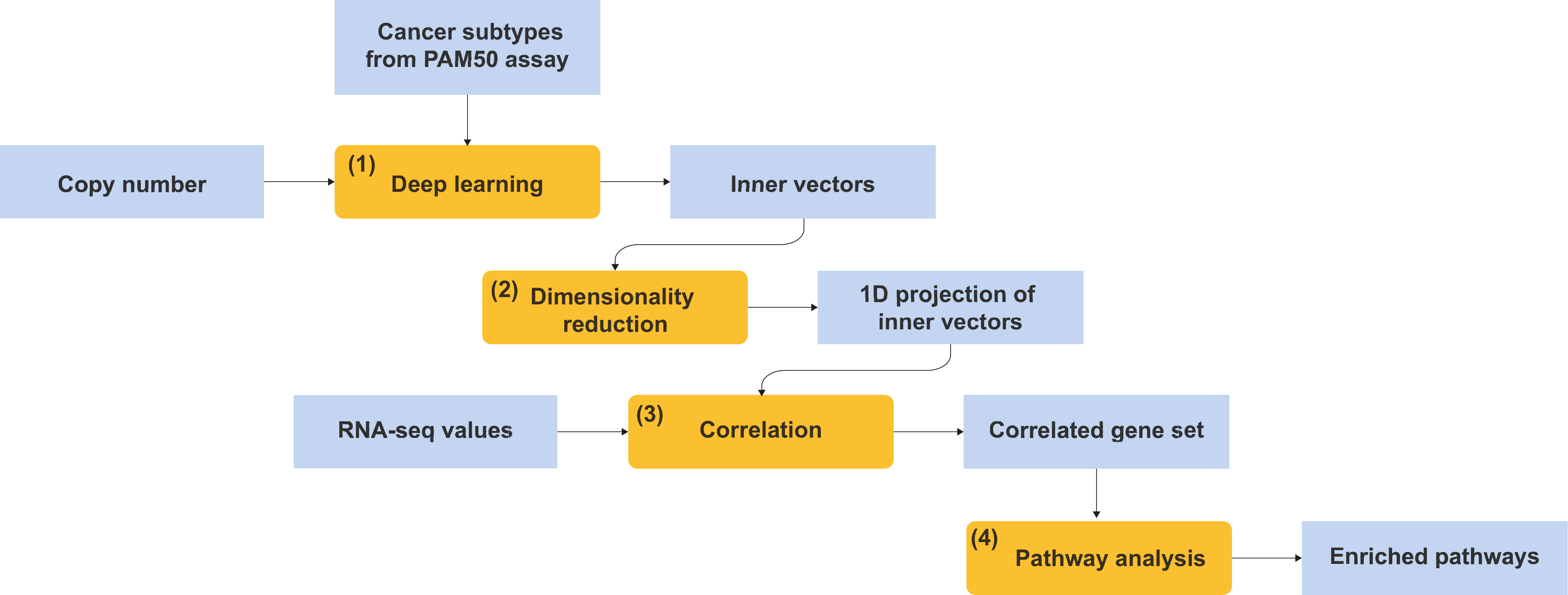}
\caption{Processing pipeline of deep enrichment analysis method. Orange boxes are functions and processes. Blue boxes are input and output data. The deep learning model (orange box (1)) is a trained deep learning model (PWL) with copy number values as the feature vectors and subtypes as the target variables. The enrichment analysis process flow is as follows. (1) The deep learning model outputs the inner vectors of the hidden layer. (2) The dimensionality reduction function projects the inner vectors to 1-dimensional (1D) variables. (3) The correlation process extracts the highly correlated gene set between RNA-seq values and the projected 1D variables. (4) The pathway analysis enriches canonical pathways from the correlated gene set.}
\label{fig:main}
\end{center}
\end{figure}

To evaluate the prediction performance of the PWL model, we prepared a classification task with RNA-seq values as the feature vectors and the five subtypes obtained from the PAM50 assay as the target variables. PAM50 was originally developed as a predictor of the five intrinsic subtypes from the expression pattern of 50 genes determined using a microarray \cite{sorlie2003repeated}. If the important genes of the PWL model with RNA-seq values include the PAM50 genes, the PWL model will be semantically validated. While PAM50 subtyping is helpful for diagnosis and stratified treatment, it remains unclear which genes contribute to the mechanisms of action and/or mechanisms of resistance to treatment for each subtype. Copy number aberrations, i.e., deletion or amplification of large continuous segments of chromosomes, are a common type of somatic mutation in cancer and can be directly associated with the expression of genes and the development of cancer \cite{cox2005survey, upender2004chromosome, burrell2013, mcgranahan2015}. Here, we newly developed a deep enrichment analysis method to investigate whether the characteristics and mechanisms of cancer were embedded in the PAM50 prediction model trained with the copy numbers. Figure \ref{fig:main} shows the processing pipeline of our deep enrichment analysis method. In the first step (Fig. \ref{fig:main} (1)), we prepared a deep learning model (PWL) with copy number values as the feature vectors and the subtypes as the target variables. In the following steps (Fig. \ref{fig:main} (2) and (3)), we calculated the correlation between the inner vector and the RNA-seq values to analyze the relationships between RNA-seq values and copy numbers. If the deep learning model can predict the subtypes of PAM50 derived from the mRNA expression level, valuable information to dictate the subtypes might be distilled in the inner vector of the deep learning model. In the final step (Fig. \ref{fig:main} (4)), we enriched canonical pathways from the highly correlated genes between RNA-seq values and copy numbers.

A paper overview (and guide for readers) is provided in \nameref{supp:overview}., where the parts of the Materials and Methods and Results sections dealing with the breast cancer subtype analysis are indicated in orange. Readers who want to grasp essential information rapidly can take a quick look at the sections and subsections indicated by check marks.

\section*{Materials and Methods}
\subsection*{Point-wise linear models}
\begin{figure}[h]
\begin{center}
\includegraphics[width=\linewidth]{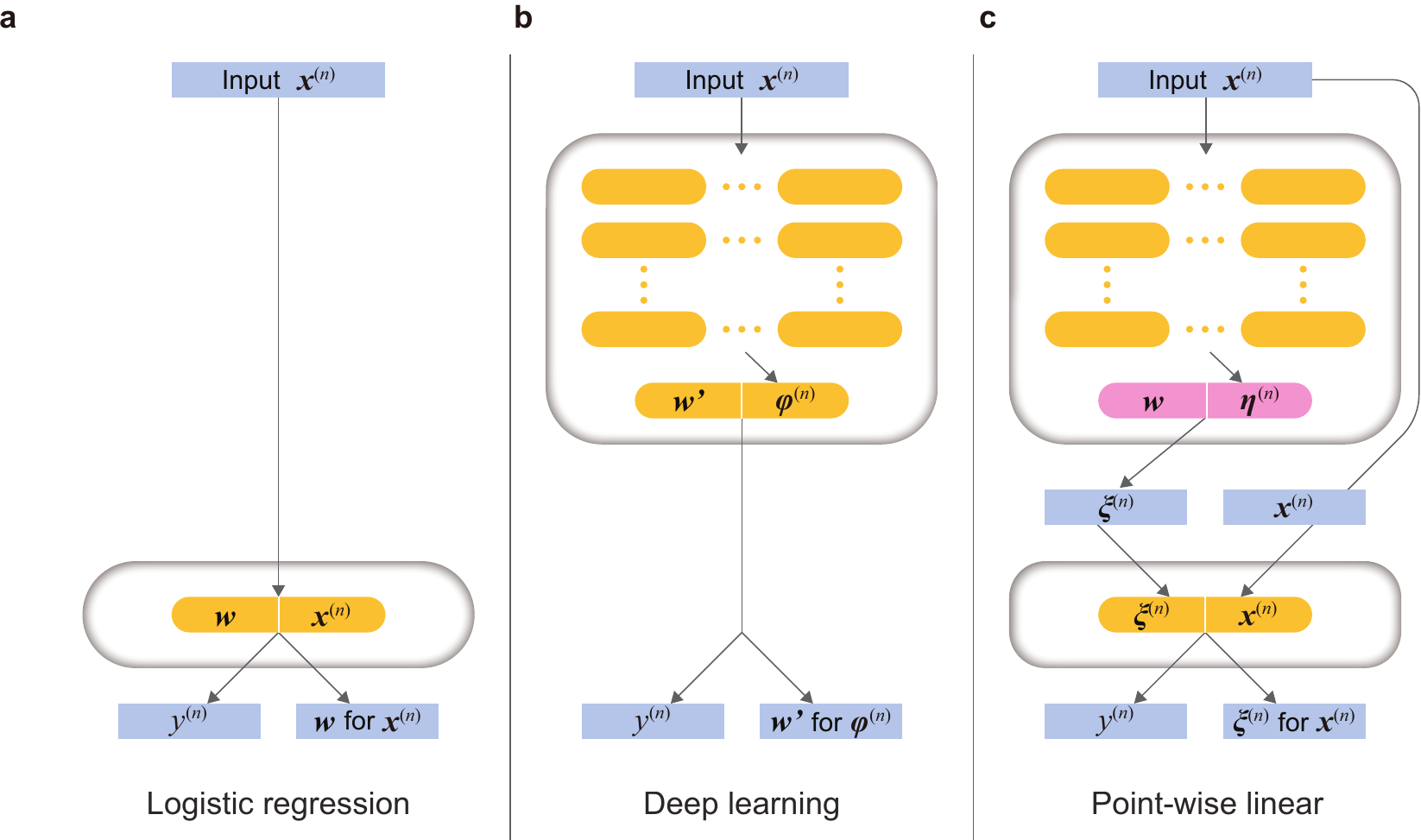}
\caption{Comparison of network architectures. (a) shows a logistic regression model. $\bm{x}^{(n)}$ and $y^{(n)}$ indicate a feature vector and a target value ($(n)$ is sample index), respectively. $\bm{w}$ is a vector of learning parameters for $\bm{x}^{(n)}$. (b) shows a fully connected neural network. $\bm{\varphi}^{(n)}$ and $\bm{w}'$ indicate an inner vector and learning parameters, respectively. (c) shows a PWL model. The upper block in (c) is a meta-machine generating a learning parameter $\bm{\xi}(\bm{x}^{(n)})$. The lower block in (c) is a logistic regression model for each feature vector $\bm{x}^{(n)}$.}
\label{fig:pwl}
\end{center}
\end{figure}

\begin{figure}[h]
\begin{center}
\includegraphics[width=\linewidth]{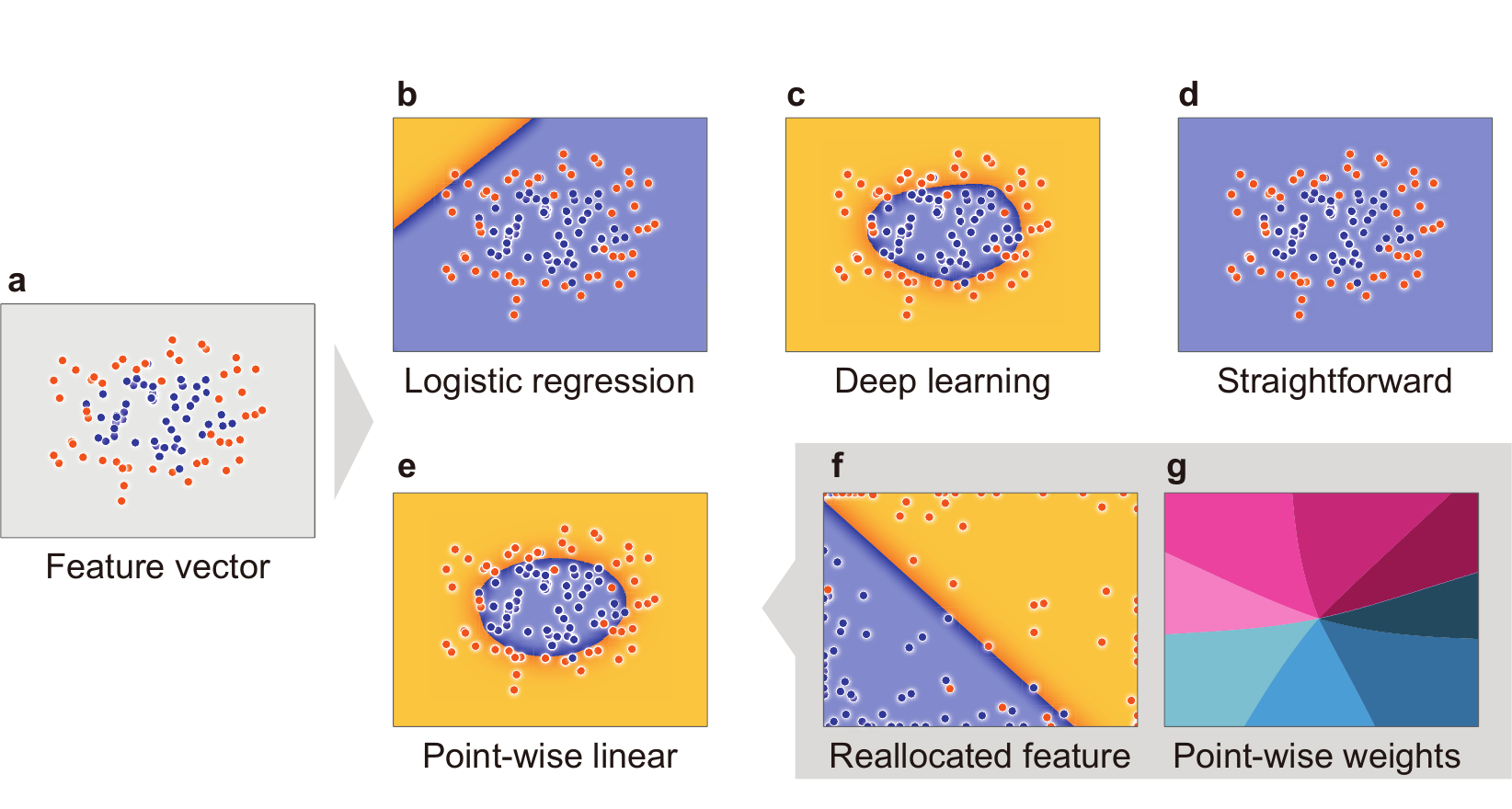}
\caption{Comparison of learning ability and explainability. (a) is a large circle (orange dots) that contains a smaller circle (blue dots) obtained by {\it sklearn.datasets.make\_circle}. (b) and (c) are the boundaries classified by the logistic regression model and self-normalizing network (SNN) model, respectively. (d) and (e) are the boundaries classified by the PWL model in a straightforward manner (Eq.~\eqref{eq:meta_classifier}) and by using the reallocation function (Eq.~\eqref{eq:point_wise_linear_inter}), respectively. (f) is the boundary classified by the reallocated feature vectors $\bm{\rho}$. (g) is the arctangent of the angle between the horizontal and vertical elements of the weight vector $\bm{\xi}^{(n)}$. The weight vector smoothly changes for each data sample.}
\label{fig:circle}
\end{center}
\end{figure}

To investigate the nonlinear prediction ability and explainability of the PWL model, we trained a logistic regression model and a self-normalizing neural network (SNN) model as a state-of-the-art deep learning \cite{DBLP:journals/corr/KlambauerUMH17} using a simple dataset (a large circle containing a smaller circle generated by {\it sklearn}\-.{\it datasets}\-.{\it make\_circle} \cite{scikit-learn}). Figure \ref{fig:pwl} shows three architectures of the machine learning models: (a) logistic regression, (b) deep learning, and (c) PWL. Let $\bm{x}^{(n)} \in \mathcal{R}^D$ represent a feature vector with $N$ denoting the sample size and $\mathcal{R}$ indicating the real number set. First, we define a logistic regression model (Fig. \ref{fig:pwl} (a)) as 

\begin{align}
 \label{eq:logit}
 y^{(n)} = \sigma(\bm{w} \cdot \bm{x}^{(n)}),
\end{align}
where $\bm{w} \in \mathcal{R}^{D}$ is a weight vector for $\bm{x}^{(n)}$, $\sigma$ is a sigmoid function, and $\cdot$ is the inner product. $y^{(n)}$ is a probability value such as one expressing the likelihood of tumor tissues or normal tissues. The weight vector $\bm{w}$ is bound to the feature vector $\bm{x}^{(n)}$. We can determine the importance of each feature variable by analyzing the magnitude of the elements in $\bm{w}$. However, as shown in Fig. \ref{fig:circle}, since the circle in a circle is not a linearly separable problem, the logistic regression model cannot classify the two circles.

Next, we define a standard deep learning model like that shown in Fig. \ref{fig:pwl} (b). A new feature vector $\bm{\varphi}(\bm{x}^{(n)}) \in \mathcal{R}^{D'}$ is nonlinearly generated from the original feature vector $\bm{x}^{(n)}$ through the $L$-layer neural network ($L \geq 2$). Note that the notation $\bm{v}\left(\bm{u}\right)$ for arbitrary vectors $\bm{v}$ and $\bm{u}$ indicates that every element of $\bm{v}$ is a function of the elements of $\bm{u}$. A deep-learning-based nonlinear classification function predicts the probability $y^{(n)}$ (Fig. \ref{fig:pwl} (b)) as follows:

\begin{align}
 \label{eq:deep_classifier}
 y^{(n)} &= \sigma\left(\bm{w'} \cdot \bm{\varphi}(\bm{x}^{(n)})\right),
\end{align}
where $\bm{w}' \in \mathcal{R}^{D'}$ is a universal weight vector for $\bm{\varphi}$. The magnitude of each $\bm{w}'$ element represents the contribution of the corresponding element of $\bm{\varphi}$ to the prediction, as shown in Fig. \ref{fig:pwl} (b). The SNNs correctly classified the blue and orange dots, as shown in Fig. \ref{fig:circle} (c). However, we cannot ``explain'' the machine's prediction by $\bm{w}'$ because it is not possible to understand the meanings of $\bm{\varphi}$ that the machine uses to make its predictions. 

In order to make a deep NN explainable, we investigated a meta-learning approach to generate a logistic regression model defined (Fig. \ref{fig:pwl} (c)) as

\begin{align}
 \label{eq:meta_classifier}
 y^{(n)} &= \sigma\left(\bm{\xi}(\bm{x}^{(n)}) \cdot \bm{x}^{(n)}\right),
\end{align}
where each element of $\bm{\xi} \in \mathcal{R}^D$ is a function of $\bm{x}^{(n)}$ that the NN determines. $\bm{\xi}$ behaves as the weight vector for the original feature vector $\bm{x}^{(n)}$. The magnitude of each element of $\bm{\xi}$ describes the importance of the corresponding feature variable. We should point out here that this weight vector is tailored to each sample because $\bm{\xi}$ depends on $\bm{x}^{(n)}$. We call Eq.~\eqref{eq:meta_classifier} a PWL model given by a straightforward method over the sample index $(n)$. The architecture of the PWL model consists of the two blocks shown in Fig. \ref{fig:pwl} (c). Also, we refer to $\bm{\xi}$ as the point-wise weight. The upper block is a meta-learning machine that generates logistic regression models, and the lower block shows the logistic regression models for the inference task. However, the tailored weight vector $\bm{\xi}$ can easily lead to poor generalization (i.e., over or underfitting). In this case, the PWL model (Eq.~\eqref{eq:meta_classifier}) tries to learn the labels of all samples because it generates a weight vector optimized for each sample, which leads to the underfitting shown in Fig. \ref{fig:circle} (c). 

We came up with a new equation that constructs a point-wise weight $\bm{\xi}$ without losing generalization ability, as follows:

\begin{align}
 \label{eq:point_wise_linear_inter}
 \bm{\xi}(\bm{x}^{(n)}) \equiv \bm{w} \odot \bm{\eta}(\bm{x}^{(n)}), 
\end{align}
where the reallocation vector $\bm{\eta} \in \mathcal{R}^{D}$ is nonlinearly generated from the original feature vector $\bm{x}^{(n)}$ through the $L$-layer NN ($L \geq 2$). $\bm{w} \in \mathcal{R}^{D}$ is a universal weight vector that is independent of $\bm{x}^{(n)}$, and $\odot$ is the Hadamard product. In contrast to the model defined straightforwardly by Eq.~\eqref{eq:meta_classifier}, the model defined by Eq.~\eqref{eq:point_wise_linear_inter} accurately predicts the classification boundary, as shown in Fig. \ref{fig:circle} (e). The weight vectors $\bm{\xi}^{(n)}$ in Eq.~\eqref{eq:point_wise_linear_inter} smoothly change for each data sample (Fig. \ref{fig:circle} (g)). The reallocation-based PWL model thus enables generalization. Additionally, we call $\bm{\rho}(\bm{x}^{(n)}) \equiv \bm{\eta}(\bm{x}^{(n)}) \odot \bm{x}^{(n)}$ a reallocated feature vector in $\mathcal{R}^d$. NNs have the versatile ability to map a linear feature space to a nonlinear feature space. By utilizing this ability, $\bm{\eta}^{(n)}$ reallocates the feature vector $\bm{x}^{(n)}$ into the new vector $\bm{\rho}$ that is linearly separable by a single hyperplane drawn by $\bm{w}$. The mechanism of Eq. \eqref{eq:point_wise_linear_inter} is discussed in \nameref{supp:appendix}.

\subsection*{Datasets}
To validate the PWL model, we used the breast cancer TCGA \cite{TCGA} dataset retrieved by the UCSC public Xena hub \cite{goldman2020visualizing} for the gene expression RNA-seq dataset (dataset ID: \-TCGA\-.BRCA\-.sampleMap\-/\-HiSeqV2\-\_PANCAN), the copy number alteration (gene-level) dataset (dataset ID: \-TCGA.BRCA.\-sampleMap\-/\-Gistic2\-\_CopyNumber \-\_Gistic2\-\_all\-\_thresholded\-.by\-\_genes), and the phenotype dataset (dataset ID: \-TCGA\-.BRCA\-.sampleMap\-/\-BRCA\-\_clinicalMatrix). RNA-seq values were calculated by UCSC Xena as follows. $Log_2(x+1)$ values were mean-normalized per-gene across all TCGA samples ($x$ is RSEM normalized count \cite{li2011rsem}). In the copy number dataset of UCSC Xena, GISTIC2 values were discretized to $-2, -1, 0, 1, 2$ by Broad Firehose. We used subtypes pre-calculated by PAM50 (Luminal A, Luminal B, basal-like, Her2-enriched, and normal-like \cite{perou2000molecular}) in the Xena dataset as the target variables of the prediction model \cite{cancer2012comprehensive}. Table \ref{tab:features} lists the number of samples for each subtype. The number $D$ (feature dimension) of mRNAs types was 17,837, as we adopted gene symbols that overlap with both the RNA-seq data and the copy number alteration data.

\begin{table}[!h]
\caption{Number of samples for each breast cancer subtype (Total = 810).}
\begin{center}
\begin{tabular}{cccccc}
& Normal-like & Luminal A & Luminal B & Basal-like & Her2-enriched \\
\hline
Number of samples & 22 & 406 & 185 & 131 & 66 \\
\end{tabular}
\end{center}
\label{tab:features}
\end{table}

\subsection*{Feature importance calculation method}
We utilized the PWL model (Eq.~\eqref{eq:meta_classifier}) to calculate the importance of each feature variable, i.e., how much each feature contributes to the model's prediction. The point-wise weight vector $\bm{\xi}$ depends on $\bm{x}^{(n)}$ and consequently describes each sample's own feature importance. Therefore, we came up with a method to derive the feature importance for a sample group so as to reveal both the group and the macroscopic property contained in the point-wise weight vector of the group's samples. This concept of the feature importance was also used in~\cite{kumagai2020pd}.

First, we calculated the feature importance for each sample from the weight vector $\bm{\xi}^{(n)}$ in Eq. \eqref{eq:meta_classifier}. Inspired by the Shapley value \cite{Shapley195317AV}, we introduced a sample-wise importance score for the $k$-th feature $x_k$ of a sample with index $(n)$ as

\begin{align}
 \label{eq:shapley_B3}
 s_k^{(n)} &\equiv \xi_k ^{(n)} x_k^{(n)} - {1 \over |U_{(n)}|}\sum_{i \in U_{(n)}}\left[ \xi_k^{(i)} x_k^{(i)} \right] ,
\end{align}
where $U_{(n)}$ is the set of samples whose weights are close to those of sample $(n)$. This sample-wise importance score expresses the contribution of a sample $(n)$ to raising the output probability $y^{(n)}$ compared with the average contribution among a sample group whose members obey similar linear models. In this study, we defined $U_{(n)}$ as follows: a sample $(i)$ is in the set $U_{(n)}$ if $ | \bm{\xi}^{(i)} - \bm{\xi}^{(n)} | / |\bm{\xi}^{(n)} |$ is smaller than $4 |\bm{\sigma}| / |\bar{\bm{\xi}}|$, where $\bm{\sigma}$ and $\bar{\bm{\xi}}$ are vectors whose elements are given as the standard deviation and the mean, respectively, of the corresponding element of $\bm{\xi}$. 

Next, we defined group-wise importance scores for each group (e.g., subtype Her2 samples) by using the sample-wise importance score. We implemented voting among the group, where each sample votes on its top $10$ \% features as to which sample-wise importance scores are the highest, i.e., the features that significantly raise each sample's output probability. We defined a group-wise importance score $v$ for each feature as the rate of samples who vote for the feature in the above voting.

Finally, we introduced a relative score to extract the features that characterize a subtype. We divided the samples into samples of a target subtype and others and then evaluated the group-wise importance score for each group. We refer to the group-wise importance score for the target subtype samples group as $v_{\rm target}$ and the one for the others as $v_{\rm others}$. $v_{\rm target}$ is not necessarily appropriate for extracting the features that characterize the target subtype because even when $v_{\rm target}$ is high for a feature, if $v_{\rm others}$ is also high, the feature might be important for all the subtype samples, not just for the target ones. Therefore, we defined a relative score $v_{\rm rel}$ so as to compute the feature importance for the subtype samples relative to the one for the others, as

\begin{align}
 \label{eq:score_rel}
 v_{{\rm rel}} &\equiv \left(v_{\rm target} \right)^2 - \left(v_{\rm others} \right)^2 .
\end{align}
Since the range of both $v_{\rm target}$ and $v_{\rm others}$ is $[0, 1]$, the range of $v_{\rm rel}$ becomes $[-1, 1]$. \nameref{supp:relative}. shows the distribution of $v_{{\rm rel}}$ in the $v_{\rm others}$-$v_{\rm target}$ space. If a relative score is large, we can expect that both the summation and difference of the group-wise importance scores $v_{\rm target}$ and $v_{\rm others}$ will be large as well. In other words, features with large relative scores are important to a certain degree for all the samples, and simultaneously they are much more important for the target subtype samples than for the others. We considered the features with high relative scores for each subtype to be the important features that characterize the corresponding subtype.

\subsection*{Deep enrichment analysis method}
In addition to the importance scores, the PWL model implements a deep learning-based enrichment analysis method to extract the canonical pathway in the nature of the breast subtypes, as summarized in Fig. \ref{fig:main}. Prior to the enrichment analysis, we trained the deep learning model (indicated by the orange box (1)) with copy number values as the feature vectors and the subtypes as the target variables. Step (1) of the pipeline utilizes the inner vectors of the hidden layer. The outputs of the NN (deep learning) inner layers give us some hints as to what criteria the NNs use to classify the subtypes of breast cancer from the feature vector. The output of the final inner layer (a new feature vector $\bm{\varphi}$) can be linearly separated by a single hyperplane spanned by $\bm{w}'$, as shown in Eq.~\eqref{eq:deep_classifier}. When the prediction accuracy is high, the output of the final inner layer provides a well-summarized representation of the feature vector $\bm{x}$ for the classification task. We utilized the reallocation vector $\bm{\eta}$ as the inner vectors, as shown in Fig. \ref{fig:main} (the output of (1)).

To determine the subtype classification criteria, the second step of the pipeline (Fig. \ref{fig:main} (2)) utilizes a manifold learning technique called UMAP for the dimensionality reduction\cite{mcinnes2018umap}. The UMAP then compresses the reallocation vectors $\bm{\eta}$ into a one-dimensional (1D) vector. The third step (Fig. \ref{fig:main} (3)) then calculates the Spearman’s correlation coefficient for the relationship between the 1D vector (the projection of $\bm{\eta}$) and RNA-seq values, after which we select the top 250 genes for which the correlation coefficient was positive and the top 250 genes for which the correlation coefficient was negative. The final step of the pipeline (Fig. \ref{fig:main} (4)) analyzes these genes by using Ingenuity Pathway Analysis (IPA, QIAGEN,\cite{kramer2014causal}) to interpret the canonical pathways.

\subsection*{Evaluation methodology}
The subtype prediction models were built by using logistic regression with regularization (implemented by scikit-learn v0.22, Python 3.7.6), SNNs (implemented by PyTorch 1.5, Python 3.7.6), and PWL (implemented by PyTorch 1.5, Python 3.7.6). Our objective was to obtain a more accurate explainable model using deep learning. Therefore, we compared the explainability of the PWL method (our proposed method) with the explainability of representative logistic regression. 

If the hyperparameters of a prediction model (e.g., the number of layers) are optimized by using all samples, we may overlook the hyperparameter overfitting. To address this issue, we carried out the prediction model evaluation by a $K$-fold double cross-validation (DCV) \cite{wang2007accurate}. The $K$-fold DCV can measure the prediction performance of the entire learning process including its hyperparameter optimization. The procedure of $K$-fold DCV has internal (training) and outer (test) loops. In this study, each internal loop searches for the best hyperparameter set (i.e., best combinations of the hyperparameters) of the prediction model $L$ times by using a tree-structured Parzen estimator \cite{bergstra2013making,optuna_2019}, where a single nested loop in the inner loops uses $M$-fold CV to evaluate the prediction performance of the prediction model with a hyperparameter set. Each inner loop trains the prediction models with different hyperparameter sets $L \times M$ times. Then, each outer loop tests the prediction model with the best hyperparameter set obtained by its internal loop. The hyperparameter optimization process is thus completely separated from the test data. In our experiment, we set M = K = 10 and L = 100, and trained the prediction model by 10,000 ($K \times L \times M$) times.

In each iteration of the 10-fold DCV and its internal 10-fold CV, the mean area under the curve (AUC) was calculated as 

\begin{align}
 \label{eq:auc}
 {\rm Mean\ AUC} = \frac{1}{KC}\sum_{k=1}^{K} \sum_{t \in subtypes} {\rm AUC}(k, t),
\end{align}
where $subtypes$ is a set of subtype categories ($C = 5$: Normal, Luminal A, Luminal B, Basal and Her2), and ${\rm AUC}(k, t)$ is the $k$-th and subtype $t$'s AUC.

According to the IPA’s help and support pages, the p-value is calculated using the right-tailed Fisher Exact Test. The p-value for a pathway is thus calculated by considering:
\begin{enumerate}
\item the number of genes that participate in that pathway, and
\item the total number of genes in the QIAGEN Knowledge Base that are known to be associated with that pathway.
\end{enumerate}

\section*{Results}
\subsection*{Prediction performance}
In advance of the breast cancer subtypes analysis, we evaluated three subtype prediction models: one built by using logistic regression with regularization, one using SNNs (state-of-the-art feed-forward NNs), and the proposed PWL. 

Table \ref{table:10foldCV} lists the AUC values of 10-fold DCV for the RNA-seq and the copy number features. All data of the 10-fold DCV was stored in \nameref{supp:10dcv}). The hyperparameter search ranges and values of each prediction model were stored in \nameref{supp:hp}. Values in the 'All' column in Table \ref{table:10foldCV} were calculated using Eq.~\eqref{eq:auc}. The AUC value of each subtype was then averaged by the 10-fold DCV. All models with the RNA-seq features achieved AUC over 0.90. The logistic regression model was better than the deep learning models. The PWL model with the copy number features marked the best values (training: 0.859, test: 0.862) of 'All'. The AUC values of the SNN model were lower than 0.8.

\begin{table}[h]
\caption{Mean AUC values of 10-fold DCV results.}
\begin{center}
\scalebox{0.90}[0.90]{
\begin{tabular}{cc|c|cccccc}
\multirow{2}{*}{Feature vectors} &\multirow{2}{*}{Models} & Training & \multicolumn{6}{c}{Test}\\
 & & All & All & Normal & Luminal A & Luminal B & Basal & Her2 \\ \hline \hline
\multirow{3}{*}{RNA-seq} & Logistic & 0.983 & 0.985 & 0.984 & 0.977 & 0.972 & 0.999 & 0.993 \\
 & SNNs & 0.933 & 0.930 & 0.843 & 0.948 & 0.930 & 0.990 & 0.940 \\
 & PWL & 0.980 & 0.975 & 0.956 & 0.966 & 0.964 & 0.999 & 0.992 \\ \hline
\multirow{3}{*}{Copy number} & Logistic & 0.851 & 0.853 & 0.883 & 0.805 & 0.746 & 0.985 & 0.845 \\
 & SNNs & 0.772 & 0.745 & 0.644 & 0.752 & 0.615 & 0.972 & 0.742 \\
 & PWL & 0.859 & 0.862 & 0.879 & 0.817 & 0.779 & 0.986 & 0.848
\end{tabular}
}
\end{center}
Logistic: Logistic regression \\
Normal: Normal-like \\
Basal: Basal-like \\
Her2: Her2-enriched
\label{table:10foldCV}
\end{table}

\subsection*{Importance analysis}
We further investigated the PWL model by comparing the overlap between the important genes and the PAM50 genes. Then, the top 500 genes in terms of the relative score (stored in \nameref{supp:imp}) calculated by the importance analysis (see Feature importance calculation in the Methods section) were selected as highly contributing feature variables to predict subtypes. We then checked how many PAM50 genes were included (\nameref{supp:pam50_top500}.). Note that the SNN model (unexplainable deep learning) cannot provide feature importance. Table \ref{table:PAM50overlap} summarizes the overlap rate between the Top 500 and the PAM50 genes. As expected, when trained with RNA-seq, both the logistic regression and PWL models contained 44 and 41 of PAM50 genes, respectively, but when training on copy number data, the overlaps of PAM50 genes were low: 14 and 15 out of 50 genes, respectively.

\begin{table}[h]
\caption{Number of overlaps between PAM50 genes and top 500 important genes in each model.}
\begin{center}
\begin{tabular}{ccccccccc}
Feature vectors            & Models    & All & Normal & Luminal A & Luminal B & Basal & Her2 \\ \hline\hline
\multirow{2}{*}{RNA-seq}  & Logistic   & 44    & 10     & 27        & 22        & 15    & 13   \\
                          & PWL        & 41    & 7      & 22        & 22        & 10    & 13   \\ \hline
\multirow{2}{*}{Copy number}
                          & Logistic   & 14    & 1      & 3         & 4         & 3     & 8    \\
                          & PWL        & 15    & 5      & 3         & 1         & 1     & 7     
\end{tabular}
\end{center}
Logistic: Logistic regression \\
Normal: Normal-like \\
Basal: Basal-like \\
Her2: Her2-enriched
\label{table:PAM50overlap}
\end{table}

\begin{figure}[h]
\begin{center}
\includegraphics[width=\linewidth]{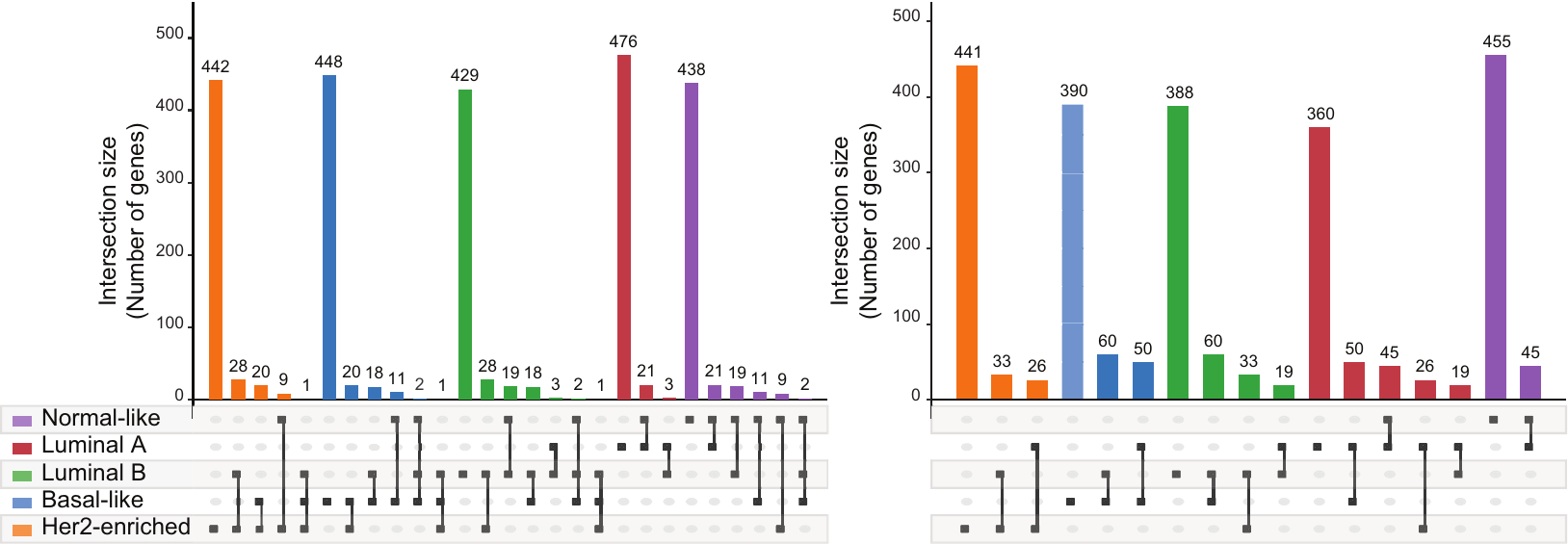}
\end{center}
\caption{Intersections of top 500 gene sets calculated by importance analysis of the copy number. (a) PWL model. (b) Logistic regression model.}
\label{fig:intsec_copynum}
\end{figure}

To examine which aspects of these contributing feature genes differed, in each model trained with copy number data, we compared those genes across the subtypes in Fig. \ref{fig:intsec_copynum} (\nameref{supp:intsec_copynum}. shows the results of the RNA-seq). We found that the majority of the contributing feature genes were specific in certain subtypes because they were not selected in other subtypes. In addition, a comparison between modeling methods showed that the PWL model tended to have a higher proportion of the specific genes than the logistic regression model for all tumor subtypes. Duplication of the specific genes between the two models showed not much commonality with 206 genes even in the most overlapping Her2 subtype, as summarized in Table \ref{table:specific} (\nameref{supp:specific} summarizes the results of the RNA-seq). The Her2 subtype is a class in which the amplification of the HER2 gene is enriched, and many genes in the vicinity of the HER2 (also known as ERBB2 and is located on chromosome 17) gene were commonly contained (please see the ideogram of chromosome 17 as shown in \nameref{supp:cytogenetic}.). \nameref{supp:chromosomal} summarizes the chromosomal location for the common genes as the specific features of the Her2-enriched class in the logistic regression and PWL models. Since both models were generated from copy number data, it is reasonable that they had a relatively high degree of commonality.

\begin{table}[!h]
\caption{Commonalities of specific feature genes for copy number.}
\label{table:specific}
\begin{center}
\begin{tabular}{ccccccc}
\multicolumn{1}{l}{} & & \multicolumn{5}{c}{PWL} \\
\multicolumn{1}{l}{} & & \multicolumn{1}{l}{Normal-like} & \multicolumn{1}{l}{Luminal A} & \multicolumn{1}{l}{Luminal B} & \multicolumn{1}{l}{Basal-like} & \multicolumn{1}{l}{Her2-enriched} \\ \hline
\multirow{5}{*}{
\rotatebox[origin=c]{90}{Logistic}
}
 & Normal-like & 37 & 21 & 0 & 0 & 0 \\
 & Luminal A & 10 & 43 & 39 & 3 & 0 \\
 & Luminal B & 4 & 0 & 117 & 7 & 4 \\
 & Basal-like & 2 & 60 & 0 & 82 & 3 \\
 & Her2-enriched & 6 & 3 & 3 & 3 & 206 \\
\end{tabular}
\end{center}
Logistic: Logistic regression
\end{table}

\subsection*{Enrichment analysis}
The previous subsection compared the PWL and logistic regression models by evaluating the overlap rates of the PAM50 genes. As Table \ref{table:PAM50overlap} shows, the PWL model with the copy number features used genes other than the PAM50 genes for subtype classification. In this subsection, we performed enrichment analysis to determine which pathways were used by the PWL model with the copy number features to classify the subtypes. Then, we compared the pathways enriched from the PWL model and the SNNs model as deep learning.

\begin{figure}[h]
\includegraphics[width=\linewidth]{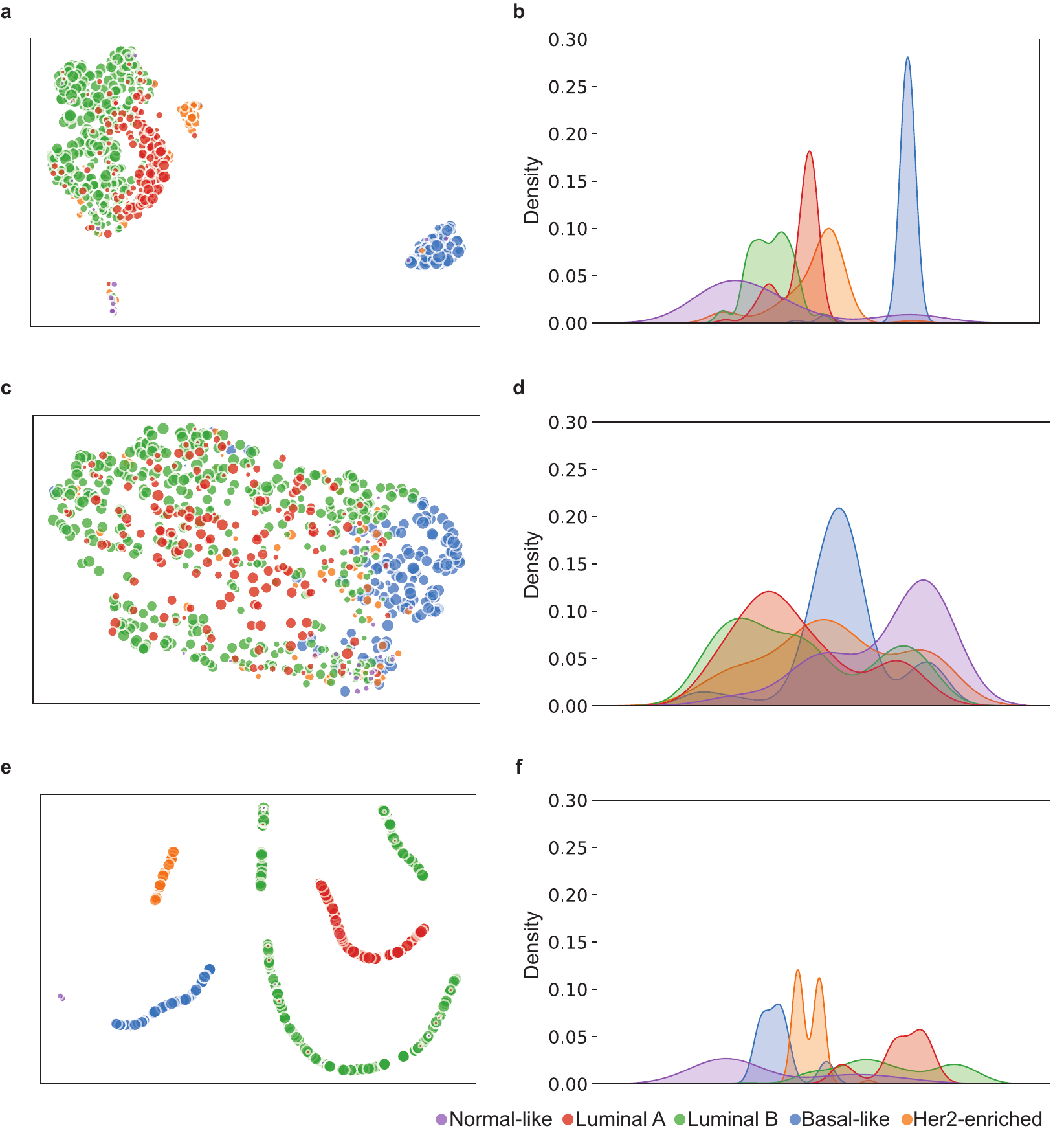}
\caption{Embeddings of deep learning models with copy number features projected by UMAP. (a), (b) Top, (c), (d) middle, and (e), (f) bottom panels are 1D and 2D embeddings of RV $\bm{\eta}$, RF $\bm{\rho}$, and SNNs, respectively.}
\label{fig:umap}
\end{figure}

Figure \ref{fig:umap} shows the 1D and 2D embeddings of the reallocation vector (RV $\bm{\eta}$) and the reallocated feature (RF $\bm{\rho}$, and the final inner vector of SNNs ($\bm{\varphi}$ in Eq.~\eqref{eq:deep_classifier}) for the copy number features (\nameref{supp:umap}. shows the results of the RNA-seq). The RV vector reflected the clinical features (ER−/PR−/Her2−). As shown in Fig. \ref{fig:umap} (a), Luminal A and B were stuck together, but each peak of their 1D embeddings stood in a distinct position. Her2-enriched samples were close to the clusters of the Luminal A and B samples. The 2D embedding of RV $\bm{\eta}$ (Fig. \ref{fig:umap} (a)) shows that the basal-like samples stayed away from other subtypes. Normal-like samples took a position under the Luminal and Her2-enriched samples. In the case of RF$\bm{\rho}$ embeddings, all subtypes formed a single cluster, as shown in Fig. \ref{fig:umap} (c). The embeddings of SNNs were separated for each subtype, as shown in Fig. \ref{fig:umap} (d).

These subtypes were originally grouped by PAM50 on the basis of the mRNA expression level. For this reason, we investigated which kind of mRNA expression levels of the genes were associated with the 1D embeddings of RV $\bm{\eta}$, RF $\bm{\rho}$, and the inner vector of the SNN model to figure out the difference of the three vectors focusing on models from only copy number data. We obtained each of the top 500 genes correlated with the mRNA expression values and then examined their functionally enriched pathways using IPA. Those genes derived from RV $\bm{\eta}$ were overlapped with cell cycle-related pathways such as ``Kinetochore metaphase signaling pathway'', ``G2/M DNA damage checkpoint regulation'', ``Estrogen-mediated S-phase entry'', ``Mitotic roles of Polo-like kinase'', and ``Cell cycle control of chromosomal replication''. In contrast, the ones derived from RF $\bm{\rho}$ showed little significant enrichment, as summarized in Table \ref{table:pathway} (the full list is available in \nameref{supp:pathways}). The gene from the SNN model that enriched pathways suggested the occurrence of a cell cycle during growth and development; however, its statistical significance was low. The cell cycle is an essential function of cell proliferation and affects the characteristics and malignancy of cancer \cite{otto2017cell}. Both the PWL and SNN models recognized those pathways in the subtype classification, and the PWL model's $\bm{\eta}$ actually presented multiple related pathways. We performed the same analysis for the PWL model generated from only RNA-seq data (the enriched pathways were stored in \nameref{supp:pathways}) and found that RV $\bm{\eta}$ genes from RNA-seq/copy number data more significantly enriched the cell cycle pathway.

\begin{table}[!h]
\caption{Enriched pathways that the top 500 genes' mRNA expression level were associated with the 1D embeddings of RV $\bm{\eta}$, RF $\bm{\rho}$, and the inner vector of SNNs in UMAP.}
\label{table:pathway}
\begin{center}
\begin{tabular}{cccc}
\multirow{2}{*}{Canonical pathways in IPA} & \multicolumn{3}{c}{$-log_{10}(p$-value$)$} \\
& RV $\bm{\eta}$ & RF $\bm{\rho}$& SNNs \\ \hline
Kinetochore metaphase signaling pathway & 20.65 & 2.05 & 5.40 \\
Cell cycle: G2/M DNA damage checkpoint regulation & 10.65 & 0.00 & 0.00 \\
Estrogen-mediated S-phase entry & 10.23 & 0.37 & 2.70 \\
Mitotic roles of Polo-like kinase & 10.12 & 0.00 & 0.39 \\
Cell cycle control of chromosomal replication & 9.97 & 0.00 & 0.00 \\
Role of CHK proteins in cell cycle checkpoint control & 8.72 & 0.93 & 0.47 \\
Cyclins and cell cycle regulation & 8.66 & 0.28 & 1.00 \\
Role of BRCA1 in DNA damage response & 6.98 & 1.06 & 0.00 \\
Cell cycle: G1/S checkpoint regulation & 6.80 & 0.00 & 1.85 \\
tRNA charging & 1.32 & 4.83 & 0.25 \\
ERBB signaling & 0.00 & 1.32 & 3.78
\end{tabular}
\end{center}
\end{table}

\section*{Discussion}
For the RNA-seq features, the logistic regression model was a better predictor of the PAM50 subtypes than the deep learning models, as summarized in \ref{table:10foldCV}. This result is reasonable because the subtypes as the target variables were detected from the RNA-seq expression pattern analysis through the PAM50 assay. The significant relationships between the RNA-seq features and the subtypes were relatively simple and suitable for the logistic regression model. On the other hand, the feature vector size and the configuration of the large dimensional feature size ($D = 17,837$) and small sample size ($n = 810$) decreased the generalization ability of the deep learning models (training: 0.980, test: 0.975). Here, a more meaningful result than the prediction accuracy was that the deep learning model used 41 genes of PAM50 to classify the subtypes, as summarized in Table \ref{table:PAM50overlap}. The reasoning was that the PWL model with the RNA-seq features uses the PAM50 genes to predict the subtypes derived from the PAM50 assay. These results demonstrated that the PWL model was a semantically valid model and encouraged us to apply the PWL model to reveal the relationships between the copy number features and the subtypes.

PAM50 subtype prediction from the copy number features is not a trivial task compared to prediction from the RNA-seq. 
The AUC values of the PWL model were better than those of the logistic regression and SNN models throughout the 10-fold DCV. 
The SNN model experienced overfitting, as demonstrated by the higher AUC values in the training results and the lower AUC values in the test results (Table \ref{table:10foldCV}). 
One possible reason for this overfitting is that the progression of learning is confined to the upper-layer parameters: i.e., there is a lack of advanced learning in the lower layers. 
Klambauer et al. showed that SNNs could use 32 layers in the conventional machine learning dataset \cite{DBLP:journals/corr/KlambauerUMH17}. 
In our task, the number of the SNN model's inner layers (average 13, minimum 10, and maximum 16, as summarized in \nameref{supp:hp}) was lower than that of the PWL model (average 19, minimum 12, and maximum 24, as summarized in \nameref{supp:hp}). 
We thus optimized the number of both models' inner layers from 10 to 25. 
As shown in Fig. \ref{fig:pwl}, the PWL model contains a deep learning block to generate the custom-made logistic regression model, so its model likely causes overfitting when the number of layers is increased. 
We used a unified architecture (see \nameref{supp:appendix} and \cite{takuma2018machine}) in this study as a newly developed deep learning architecture characterized by the binding of each network layer, with neurons arranged in a mesh-like form. 
This unified architecture has horizontally shallow and vertically deep layers to prevent gradient vanishing and explosion. 
No matter how many layers are stacked vertically, there are only two horizontal layers from the data unit neurons to the output neurons.

The logistic regression model and the PWL model had almost equivalent prediction performances for the copy number features. On the other hand, Table \ref{table:specific} suggests that the genes important for predicting the breast cancer subtypes were different in the two models: namely, the PWL model tended to select specific genes for each subtype, as shown in Fig. \ref{fig:intsec_copynum}. This difference is considered to stem from the models' ability to treat the nonlinear relationships between feature and target variables. The logistic regression expresses the target variables only as linear combinations of the feature variables, while the PWL can express the target variables nonlinearly, as in Eqs.~\eqref{eq:meta_classifier} and \eqref{eq:point_wise_linear_inter}. In addition, these two equations can be modified as $\bm{w} \cdot (\bm{\eta}(\bm{x}) \odot \bm{x})$. The RV $\bm{\eta}$ corrects the feature variables $\bm{x}$ so that the universal weight $\bm{w}$ can linearly separate the feature vector $\bm{x}$. The differences of important genes between the PWL and logistic regression methods stem from the feature variables' correction mechanism of the PWL method. This mechanism might also help the PWL model express the target variables without using some of the features not specific to the subtype. In contrast, the logistic regression model has to use nonspecific features to express the target variables without the corrections. We expect the PWL model's correction mechanism to be vital in tasks that demand high nonlinearity and consequently result in low AUC values in logistic regression. One of the tasks in this study was to predict subtypes Luminal A and B (see Table \ref{table:10foldCV}). Consistently, the difference of the number of specific genes in the two models was large in Luminal A and B, and the specific genes for Luminal A and B in the PWL model had much in common with the specific genes of other subtypes in the logistic regression model, as shown in Table \ref{table:specific}.

As discussed above, the results in Fig. \ref{fig:intsec_copynum}, Tables \ref{table:10foldCV}, and \ref{table:specific}) are consistent with the interpretation of the PWL model from the viewpoint of the correction to the features. This consistency suggests that our scoring method (described in 'Feature importance calculation method' in the Methods section) works properly. Our scoring method was designed to extract the features that contribute to predicting a subtype, especially among the corresponding subtype samples, rather than features not found in the other samples. This scoring method helps us clarify which features contribute to the prediction result with the aid of corrections by RV $\bm{\eta}$, but we cannot examine which features contribute to the corrections $\bm{\eta}$. Therefore, the relative score is not a perfect measure to investigate the mechanism of the classification of breast cancer subtypes.

For further investigation, we analyzed our predictive model's internal state in detail by using the deep enrichment analysis method, as shown in Fig. \ref{fig:main}. The results suggest that the gene sets' biological implications contribute to the classification. From the comparison of RV $\bm{\eta}$, RF $\bm{\rho}$ as the PWL model's inner vectors to be analyzed, we found that RV $\bm{\eta}$ was better suited to elicit candidate hypotheses. As shown in Fig. \ref{fig:main} (2) through (4), we utilized UMAP to analyze the internal state in detail and reduce the dimensionality and then combined it with mRNA expression levels, which derived the biological implication that $\bm{\eta}$ was involved in the cell cycle. The fact that genes involved in the cell cycle affect subtypes is well-known \cite{breastcancer}, and such genes have been reported as promising drug targets \cite{otto2017cell}, indicating that the internal state of our model was worth analyzing. Note that the UMAP embedding of RV $\bm{\eta}$, rather than the ones of RF $\bm{\rho}$ and $\bm{\varphi}$, contains richer information related to interpretable pathways. While RF $\bm{\rho}$ is created as new features with which the problem is linearly separable in the PWL model (as well as $\bm{\varphi}$ in the SNN model), RV $\bm{\eta}$ is considered as the corrections to the features by which the features are transformed into RF $\bm{\rho}$, and is unique to the PWL model. This finding suggests that it is essential to analyze the corrections to the features for our task in order to investigate the breast cancer subtypes classification mechanism. Regarding further analysis or hypotheses, the PWL model's $\bm{\eta}$ is preferred, as it presented multiple related pathways in this report. Considering this case as an example, we feel confident that biologists and informaticians can apply our analysis with RV $\bm{\eta}$ to other tasks, such as exploring new drug target molecules or investigating mechanisms.

The limitation of this study is the retrospective analysis of the breast cancer data. All results obtained from the prediction models should be evaluated by clinical practice as a prospective study. Even so, the results of this retrospective study demonstrate the potential of our technique to reveal unknown and nonstandard knowledge of breast cancer.

\section*{Conclusion}
This study has established the PWL model as an innately explainable deep learning model that can analyze the biological mechanisms of breast cancer subtypes. The PWL model generates a custom-made logistic regression model, which allows us to analyze which genes are important for each subtype of an individual patient. We presented a new scoring method for selecting genes for the subtype classification, and also demonstrated that the deep enrichment analysis method with the PWL model can extract the genes relevant to cell cycle-related pathways. Our PWL model utilized as explainable deep learning can reveal the mechanisms underlying cancers and thereby contribute to improving overall clinical outcomes.

\section*{Author Contributions}
\paragraph*{Conceptualization:} Takuma Shibahara, Chisa Wada, Yasuho Yamashita, Kazuhiro Fujita, Masamichi Sato, Atsushi Okamoto, Yoshimasa Ono.
\paragraph*{Methodology:} Takuma Shibahara, Chisa Wada, Yasuho Yamashita, Kazuhiro Fujita.
\paragraph*{Data curation:} Chisa Wada, Kazuhiro Fujita.
\paragraph*{Software:} Takuma Shibahara, Yasuho Yamashita.
\paragraph*{Visualization:} Junichi Kuwata, Takuma Shibahara.
\paragraph*{Project administration:} Takuma Shibahara, Atsushi Okamoto, Yoshimasa Ono.
\paragraph*{Writing – original draft preparation:} Takuma Shibahara, Chisa Wada, Yasuho Yamashita, Kazuhiro Fujita.
\paragraph*{Writing – review \& editing:} Takuma Shibahara, Chisa Wada, Yasuho Yamashita, Kazuhiro Fujita, Masamichi Sato, Junichi Kuwata, Atsushi Okamoto, Yoshimasa Ono.

\section*{Supporting information}

All supporting information files are available at https://doi.org/10.5281/zenodo.6859285 .

\paragraph*{S1 Fig.} \label{supp:overview}{\bf Paper overview.} The orange background sections show the methods and results of the breast cancer subtypes analysis. Check-marked sections provide essential information about our paper. (EPS)

\paragraph*{S2 Fig.} \label{supp:relative}{\bf Heat map visualization of distribution of relative score $\bm{v}_{rel}$.} Vertical and horizontal axes represent $\bm{v}_{target}$ and $\bm{v}_{others}$, the group-wise importance scores for the target subtype samples group and the other samples group, respectively. (EPS)

\paragraph*{S3 Fig.} \label{supp:intsec_copynum}{\bf Intersections of top 500 gene sets calculated by importance analysis of RNA-seq.} (a) PWL model and (b) logistic regression model. (EPS)

\paragraph*{S4 Fig.} \label{supp:umap} {\bf Embeddings of deep learning models with RNA-seq features projected by UMAP.} (a), (b) Top, (c), (d) middle, and (e), (f) bottom panels are 1D and 2D embeddings of RV $\bm{\eta}$, RF $\bm{\rho}$, and SNNs, respectively. (EPS)

\paragraph*{S5 Fig.} \label{supp:pam50_top500} {\bf PAM50 genes selected as top 500 gene sets in subtypes.} Details of PAM50 genes selected for top 500 genes in the deep learning models are shown. (TIFF) 

\paragraph*{S6 Fig.} \label{supp:cytogenetic} {\bf Ideogram of chromosome 17, Cytogenetic.} GRCh37\.p12 (GCF\_000001405.24) (EPS)

\paragraph*{S1 Table.} \label{supp:specific} {\bf Intersections of top 500 gene sets of RNA-seq.} This table summarizes specific genes (not selected in other subtypes) extracted by PWL model and logistic regression model with RNA-seq features. (XLSX)

\paragraph*{S2 Table.} \label{supp:chromosomal} {\bf Chromosomal location for common genes as specific features of Her2-enriched class in the logistic regression and PWL models.} (DOCX)

\paragraph*{S1 Appendix.} \label{supp:appendix} {\bf Details of PWL model.} This appendix summarizes why conventional deep learning models are not explainable and then presents the detail of the PWL model, our explainable deep learning model. (PDF)

\paragraph*{S1 File.} \label{supp:10dcv}{\bf 10-fold DCV results.} This file stores the training and test AUC values of 10-fold DCV results yielded from evaluation with each technique. (XLSX)

\paragraph*{S2 File.} \label{supp:hp}{\bf Hyperparameter optimization results.} This file stores the hyperparameter search ranges and their optimized values of each prediction model. (XLSX)

\paragraph*{S3 File.} \label{supp:imp}{\bf Relative scores.} This compressed file stores the relative scores of the PWL and logistic regression models. (ZIP)

\paragraph*{S4 File.} \label{supp:pathways} {\bf Enriched pathways of deep learning models.} This file stores the full list of Table \ref{table:pathway} for the copy number features and RNA-seq features. (XLSX)

\end{document}